\documentstyle{article}
\title{Recovery of  structure of looped    
jointed objects  from  multiframes}
\author{Mieczyslaw A. K{\l}opotek\\
{\footnotesize\sl Institute of Computer Science, 
Polish Academy of Sciences}\\
{\footnotesize\sl 
e-mail:
klopotek{@}wars.ipipan.waw.pl} }

\date{}
\newcommand{\VV}[1]{{\bf #1}}
\newcommand{\len}[1]{||\VV{#1}||}

\newcommand{\V}[1]{$\VV{#1}$}
\newcommand{\Bem}[1]{}
\newcommand{\nn}{\nonumber}

\begin{document}

\setcounter{page}{645}
\pagestyle{myheadings}
\maketitle

{
\footnotesize
{\bf Abstract.}\ \
A method  to recover 
 structural parameters of looped jointed objects from multiframes 
is being developed. Each
rigid part of the jointed body needs only to be traced at two (that is at
junction) points.
 This method has been linearized for 4-part loops, with recovery from at least
19 frames.}

\section{Introduction}
\label{reviewBiege} 

In his papers  Johansson,   \cite{Johansson:1}, drew attention to the
capability
of a man to reconstruct motion and shape of an object if only two traceable
points are available on each rigid part of a jointed object.  
Most of known algorithm for recovery of structure from motion
require more (or even significantly more) traceable features to recover
motion of rigid bodies alone (compare  
\cite{Ullman:1,Mitiche:2,Mitiche:1,Lee:1,Klopotek:92g,%
Weng:1}).

Several attempts have been therefore made to make use of the jointedness
property of jointed objects to utilize the psychological observation of
Johansson.  
Clocksin, \cite{Clocksin:1}, tried to recover the structure of joins of an
object (the join structure shows which visible points are rigidly connected);
He used a heuristic approach based on prediction of motion of rigidly
connected points. First, he assumed that all points are rigidly connected and
then broke rigid connection for points not fitting the predicted pattern of
motion.

Rashid, \cite{Rashid:1}, tried also to recover heuristically the structure of
joins. He estimated relative positions and speeds of all pairs of points for a
given number of projections (25 or 30). Then he discovered the join structure
finding minimal spanning tree of the graph of all points calculating weights
of branches from correlation of position and speed of points. A weakness of
this approach is missing three dimensional interpretation of motion of points
which may lead under some circumstances to erroneous 
interpretation of joins.

To interpret Johnson's figures, 
O'Rourke and Badler, \cite{ORourke:1}, used
 background knowledge 
about the join structure of the observed jointed object 
(human body shape in their case). The model was confronted with the image
enabling to reconstruct the position even of invisible parts of the object.

Hoffman and  Flinchbaugh, \cite{Hoffman:1}, elaborated a method of
reconstruction of structure of joins and of 3-dimensional structure of 
Johansson's figures assuming that the motion of all rigid parts in the image
is planar.

Webb and Aggarwal, \cite{Webb:1,Webb:2,Webb:3}, made also restrictive
assumptions
- that all traceable points of the object rotate around a fixed rotation
 axis. 

Lee, \cite{Lee:1}, showed that one can recover structure and motion parameters
of two rigidly connected points even assuming only a fixed direction of
rotation.

\unitlength=3cm
\begin{figure}
\special{em:linewidth 0.4pt}
\begin{picture}(1.73,1.73)

\special{em:linewidth 0.2pt}
\put(1.188,0.610){\special{em:moveto} }
\put(1.238,1.539){\special{em:lineto} }
\put(1.188,0.610){\special{em:moveto} }
\put(1.454,0.244){\special{em:lineto} }
\put(1.188,0.610){\special{em:moveto} }
\put(0.225,0.557){\special{em:lineto} }
\special{em:linewidth 1.0pt}
\put(1.505,1.173){\special{em:moveto} }
\put(1.454,0.244){\special{em:lineto} }
\put(1.505,1.173){\special{em:moveto} }
\put(1.238,1.539){\special{em:lineto} }
\put(1.505,1.173){\special{em:moveto} }
\put(0.542,1.120){\special{em:lineto} }
\put(0.276,1.486){\special{em:moveto} }
\put(0.225,0.557){\special{em:lineto} }
\put(0.276,1.486){\special{em:moveto} }
\put(0.542,1.120){\special{em:lineto} }
\put(0.276,1.486){\special{em:moveto} }
\put(1.238,1.539){\special{em:lineto} }
\put(0.492,0.191){\special{em:moveto} }
\put(0.542,1.120){\special{em:lineto} }
\put(0.492,0.191){\special{em:moveto} }
\put(0.225,0.557){\special{em:lineto} }
\put(0.492,0.191){\special{em:moveto} }
\put(1.454,0.244){\special{em:lineto} }
\put(0.547,0.686){$x$}
\put(0.439,1.333){$y$}
\put(1.053,1.177){$z$}
\special{em:linewidth 0.4pt}

\put(0.880,0.514){\special{em:moveto} }
\put(1.320,0.759){\special{em:lineto} }

\put(0.988,0.752){\special{em:moveto} }
\put(1.320,0.759){\special{em:lineto} }

\put(0.988,0.752){\special{em:moveto} }
\put(0.981,0.781){\special{em:lineto} }

\put(1.328,0.479){\special{em:moveto} }
\put(0.981,0.781){\special{em:lineto} }

\put(1.328,0.479){\special{em:moveto} }
\put(0.340,0.637){\special{em:lineto} }

\put(1.060,1.259){\special{em:moveto} }
\put(0.340,0.637){\special{em:lineto} }

\put(1.060,1.259){\special{em:moveto} }
\put(0.906,0.815){\special{em:lineto} }

\put(1.036,0.668){\special{em:moveto} }
\put(0.906,0.815){\special{em:lineto} }

\put(1.036,0.668){\special{em:moveto} }
\put(0.880,0.514){\special{em:lineto} }
\end{picture}
%
\begin{picture}(1.73,1.73)

\special{em:linewidth 0.2pt}
\put(1.188,1.234){\special{em:moveto} }
\put(1.238,1.619){\special{em:lineto} }
\put(1.188,1.234){\special{em:moveto} }
\put(1.454,0.339){\special{em:lineto} }
\put(1.188,1.234){\special{em:moveto} }
\put(0.225,1.006){\special{em:lineto} }
\special{em:linewidth 1.0pt}
\put(1.505,0.724){\special{em:moveto} }
\put(1.454,0.339){\special{em:lineto} }
\put(1.505,0.724){\special{em:moveto} }
\put(1.238,1.619){\special{em:lineto} }
\put(1.505,0.724){\special{em:moveto} }
\put(0.542,0.496){\special{em:lineto} }
\put(0.276,1.391){\special{em:moveto} }
\put(0.225,1.006){\special{em:lineto} }
\put(0.276,1.391){\special{em:moveto} }
\put(0.542,0.496){\special{em:lineto} }
\put(0.276,1.391){\special{em:moveto} }
\put(1.238,1.619){\special{em:lineto} }
\put(0.492,0.111){\special{em:moveto} }
\put(0.542,0.496){\special{em:lineto} }
\put(0.492,0.111){\special{em:moveto} }
\put(0.225,1.006){\special{em:lineto} }
\put(0.492,0.111){\special{em:moveto} }
\put(1.454,0.339){\special{em:lineto} }
\put(0.547,0.334){$x$}
\put(0.439,0.974){$y$}
\put(1.053,0.640){$z$}
\special{em:linewidth 0.4pt}

\put(0.880,0.767){\special{em:moveto} }
\put(1.320,0.840){\special{em:lineto} }

\put(0.988,0.886){\special{em:moveto} }
\put(1.320,0.840){\special{em:lineto} }

\put(0.988,0.886){\special{em:moveto} }
\put(0.981,1.057){\special{em:lineto} }

\put(1.328,0.528){\special{em:moveto} }
\put(0.981,1.057){\special{em:lineto} }

\put(1.328,0.528){\special{em:moveto} }
\put(0.340,1.022){\special{em:lineto} }

\put(1.060,1.387){\special{em:moveto} }
\put(0.340,1.022){\special{em:lineto} }

\put(1.060,1.387){\special{em:moveto} }
\put(0.906,1.135){\special{em:lineto} }

\put(1.036,0.581){\special{em:moveto} }
\put(0.906,1.135){\special{em:lineto} }

\put(1.036,0.581){\special{em:moveto} }
\put(0.880,0.767){\special{em:lineto} }
\end{picture}
\caption{Two views of a complex looped
jointed object (no motion between images).
} \label{figeins}
\end{figure}
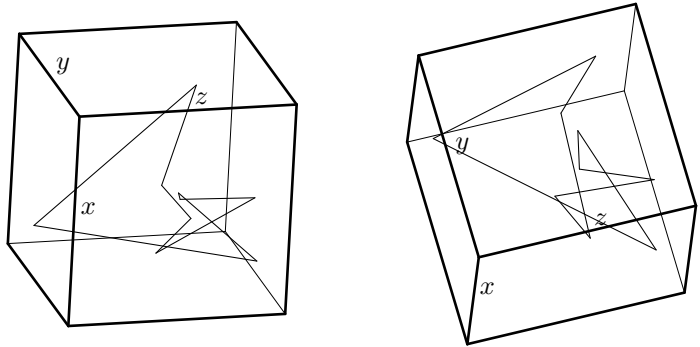

%

In this paper we investigate a different assumption about the jointed object.
We assume that the rigid parts of the object form a loop or loops (see
Fig.\ref{figeins}) and that the traceable points are the junction points.
We will concentrate on loops composed of 4 rigid parts only and demonstrate
that the task can be linearized if 19 images are available. We handle
orthogonal projections only. Planarity of motion is not required.

\section{The Method}

The fundamental approach to the problem of structure and motion from
multiframes is  to  find  and  utilize  one  (or  more)  invariant 
properties combining
visible quantities with global unknowns (not engaging unknowns local to the
single image). Surprisingly, for orthogonal projections of looped jointed
objects the property is analogous to three-point rigid objects  case 
described in
\cite{Klopotek:92g}.

Let us consider a loop consisting of points $P_1,P_2,...,P_n,P_1$. Then
obviously the sum of  vectors  \V{P_1P_2},  \V{P_2P_3}..\V{P_nP_1} 
fulfills the
equation:
\begin{equation}
    \VV{P_1P_2}+ \VV{P_2P_3}+...+\VV{P_{n-1}P_n}+\VV{P_nP_1}=\VV{0}.
\end{equation}

Let us assume that the image plane is the XY plane (the projection direction
is then along the Z-axis). The above relation is also true for component
vectors of 
these vectors in each direction. Let $\VV{P_kP_j}_z$ be the
Z-direction component of $\VV{P_kP_j}$. We have:
\begin{equation}
    \VV{P_1P_2}_z+ \VV{P_2P_3}_z+...+\VV{P_{n-1}P_n}_z+\VV{P_nP_1}_z=\VV{0}.
\end{equation}

But this means that if  $||\VV{P_kP_j}||$ denotes the length of the vector 
 $\VV{P_kP_j}$ then there exists a combination of +'s and -'s such that we get
a result equal to 0 in the equation below:
\begin{equation}
  ||\VV{P_1P_2}_z||\pm  ||\VV{P_2P_3}_z||\pm \dots
\pm ||\VV{P_{n-1}P_n}_z||\pm ||\VV{P_nP_1}_z||= 0.
\end{equation}
\noindent

Let us denote by $P_j'$ the projection of point $P_j$. Obviously:\\
$$||\VV{P_kP_j}_z|| = \sqrt{ \len{P_kP_j}^2-\len{P_k'P_j'}^2}.$$

Hence 
\begin{eqnarray}
 \sqrt{ \len{P_1P_2}^2-\len{P_1'P_2'}^2} \pm
 \sqrt{ \len{P_2P_3}^2-\len{P_2'P_3'}^2} \pm
\dots \nn\\
 \sqrt{ \len{P_{n-1}P_n}^2-\len{P_{n-1}'P_n'}^2} \pm
 \sqrt{ \len{P_nP_1}^2-\len{P_n'P_1'}^2} = 0.
\label{mostgeneral}
\end{eqnarray}

This is an equation in $n$ unknowns (lengths of line segments  $P_1P_2$,
$P_2P_3$, \dots, $P_{n-1}P_n$, $P_nP_1$). From $n$ images we can get in
principle $n$ (most probably independent) equations
and just find the unknown lengths of rigid edges of the jointed objects. Once
these lengths are known, the position of the object in space for each frame is
known (up to shift along the Z axis and reflections about the XY plane). 

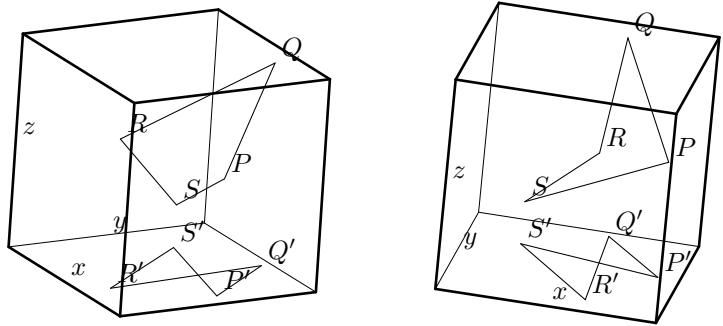
\begin{figure}
\special{em:linewidth 0.4pt}
\begin{picture}(1.73,1.73)

\special{em:linewidth 0.2pt}
\put(1.020,0.599){\special{em:moveto} }
\put(1.514,0.292){\special{em:lineto} }
\put(1.020,0.599){\special{em:moveto} }
\put(0.153,0.493){\special{em:lineto} }
\put(1.020,0.599){\special{em:moveto} }
\put(1.083,1.545){\special{em:lineto} }
\special{em:linewidth 1.0pt}
\put(0.647,0.185){\special{em:moveto} }
\put(0.153,0.493){\special{em:lineto} }
\put(0.647,0.185){\special{em:moveto} }
\put(1.514,0.292){\special{em:lineto} }
\put(0.647,0.185){\special{em:moveto} }
\put(0.710,1.131){\special{em:lineto} }
\put(1.577,1.237){\special{em:moveto} }
\put(1.083,1.545){\special{em:lineto} }
\put(1.577,1.237){\special{em:moveto} }
\put(0.710,1.131){\special{em:lineto} }
\put(1.577,1.237){\special{em:moveto} }
\put(1.514,0.292){\special{em:lineto} }
\put(0.216,1.438){\special{em:moveto} }
\put(0.710,1.131){\special{em:lineto} }
\put(0.216,1.438){\special{em:moveto} }
\put(1.083,1.545){\special{em:lineto} }
\put(0.216,1.438){\special{em:moveto} }
\put(0.153,0.493){\special{em:lineto} }
\put(0.430,0.369){$x$}
\put(0.616,0.576){$y$}
\put(0.214,0.995){$z$}
\special{em:linewidth 0.4pt}

\put(1.109,0.795){\special{em:moveto} }
\put(1.333,1.308){\special{em:lineto} }

\put(0.648,0.971){\special{em:moveto} }
\put(1.333,1.308){\special{em:lineto} }

\put(0.648,0.971){\special{em:moveto} }
\put(0.896,0.679){\special{em:lineto} }

\put(1.109,0.795){\special{em:moveto} }
\put(0.896,0.679){\special{em:lineto} }

\put(1.075,0.275){\special{em:moveto} }
\put(1.273,0.409){\special{em:lineto} }

\put(0.604,0.309){\special{em:moveto} }
\put(1.273,0.409){\special{em:lineto} }

\put(0.604,0.309){\special{em:moveto} }
\put(0.883,0.490){\special{em:lineto} }

\put(1.075,0.275){\special{em:moveto} }
\put(0.883,0.490){\special{em:lineto} }

\put(1.139,0.825){$P$}

\put(1.363,1.338){$Q$}

\put(0.678,1.001){$R$}

\put(0.926,0.709){$S$}

\put(1.105,0.305){$P'$}

\put(1.303,0.439){$Q'$}

\put(0.634,0.339){$R'$}

\put(0.913,0.520){$S'$}
\end{picture}
%
\special{em:linewidth 0.4pt}
\begin{picture}(1.73,1.73)
%
\special{em:linewidth 0.2pt}
\put(0.427,0.647){\special{em:moveto} }
\put(1.404,0.495){\special{em:lineto} }
\put(0.427,0.647){\special{em:moveto} }
\put(0.236,0.307){\special{em:lineto} }
\put(0.427,0.647){\special{em:moveto} }
\put(0.517,1.575){\special{em:lineto} }
\special{em:linewidth 1.0pt}
\put(1.213,0.155){\special{em:moveto} }
\put(0.236,0.307){\special{em:lineto} }
\put(1.213,0.155){\special{em:moveto} }
\put(1.404,0.495){\special{em:lineto} }
\put(1.213,0.155){\special{em:moveto} }
\put(1.303,1.083){\special{em:lineto} }
\put(1.494,1.423){\special{em:moveto} }
\put(0.517,1.575){\special{em:lineto} }
\put(1.494,1.423){\special{em:moveto} }
\put(1.303,1.083){\special{em:lineto} }
\put(1.494,1.423){\special{em:moveto} }
\put(1.404,0.495){\special{em:lineto} }
\put(0.326,1.235){\special{em:moveto} }
\put(1.303,1.083){\special{em:lineto} }
\put(0.326,1.235){\special{em:moveto} }
\put(0.517,1.575){\special{em:lineto} }
\put(0.326,1.235){\special{em:moveto} }
\put(0.236,0.307){\special{em:lineto} }
\put(0.754,0.261){$x$}
\put(0.361,0.507){$y$}
\put(0.311,0.801){$z$}
\special{em:linewidth 0.4pt}

\put(1.270,0.868){\special{em:moveto} }
\put(1.089,1.420){\special{em:lineto} }

\put(0.963,0.909){\special{em:moveto} }
\put(1.089,1.420){\special{em:lineto} }

\put(0.963,0.909){\special{em:moveto} }
\put(0.632,0.693){\special{em:lineto} }

\put(1.270,0.868){\special{em:moveto} }
\put(0.632,0.693){\special{em:lineto} }

\put(1.221,0.357){\special{em:moveto} }
\put(1.004,0.539){\special{em:lineto} }

\put(0.900,0.259){\special{em:moveto} }
\put(1.004,0.539){\special{em:lineto} }

\put(0.900,0.259){\special{em:moveto} }
\put(0.614,0.507){\special{em:lineto} }

\put(1.221,0.357){\special{em:moveto} }
\put(0.614,0.507){\special{em:lineto} }

\put(1.300,0.898){$P$}

\put(1.119,1.450){$Q$}

\put(0.993,0.939){$R$}

\put(0.662,0.723){$S$}

\put(1.251,0.387){$P'$}

\put(1.034,0.569){$Q'$}

\put(0.930,0.289){$R'$}

\put(0.644,0.537){$S'$}
\end{picture}
\caption{Two views of a four part
jointed object PQRS (no motion between images).} \label{figzwei}
\end{figure}

Let us restrict now our treatment to  the case of only four traceable points
(a loop of
four rigid parts). Let us call the traceable points $P,Q,R,S$ and their
projections  $P',Q',R',S'$   respectively (see fig.\ref{figzwei}).  Let us
also introduce the following notation for global unknowns:\\
\begin{eqnarray}
a = {PQ}^2, 
b = {QR}^2, 
c = {RS}^2, 
d = {SP}^2.
\end{eqnarray}

Quantities obtainable (measurable) from $i$'th frame are as follows: 
\begin{eqnarray}
A_i = {P'Q'}^2, 
B_i = {Q'R'}^2, 
C_i = {R'S'}^2, 
D_i = {S'P'}^2 .
\end{eqnarray}

We will occasionally drop the $i$-index if we are talking about only one 
(current) frame. Under this notation the eq.~\ref{mostgeneral}
turns to:

\begin{equation} \label{exprX}
\sqrt{a-A_i} \pm \sqrt{b-B_i} \pm \sqrt{c-C_i} \pm \sqrt{d-D_i} = 0.
\end{equation}

We will show now  how this equation can be turned to "linear" form.
Details are given in the Appendix A. The principle is to square out
the square roots to obtain a polynomial in our four unknowns $a,b,c$ and $d$.
Thereafter auxiliary variables $x_{1}-x_{19}$ are introduced which are
themselves polynomials in $a,b,c,d$ and independent of the current frame.
Variables $x_1,x_2,x_3,x_4$ are identical with $a,b,c,d$ resp. and are of
primary interest for us (the lengths of links in the loop are then obtained as
square roots of  $x_1,x_2,x_3,x_4$).
If we represent constant expressions obtained as
free expressions and as  factors for  $x_{1}-x_{19}$ 
 as  $f_{i,0}-f_{i,
19}$ ($i$ - the index
of the frame), 
 we can write the above eq.~\ref{exprX} in the plain form: 
\begin{eqnarray} 
&&f_{i,1} x_1 + f_{i,2} x_2 + f_{i,3} x_3 + f_{i,4} x_4 + f_{i,5} x_5 
+f_{i,6} x_6 + f_{i,7} x_7 + f_{i,8} x_8 + f_{i,9} x_9 \nn\\
&&+ f_{i,10} x_{10}  
+f_{i,11} x_{11}+f_{i,12} x_{12} 
+f_{i,13} x_{13}+f_{i,14} x_{14} 
+f_{i,15} x_{15} \nn\\
&&+f_{i,16}x_{16}+
f_{i,17}x_{17}+f_{i,18}x_{18}+f_{i,19}x_{19}+ f_{i,0} =0.
\end{eqnarray}

We have now one linear equation in 19 variables for each frame. Though the
variables and coefficients are dependent, they are not linearly dependent.
Hence 19 frames from a free motion of this body may allow us to recover the
object parameters $a,b,c,d$. The subsequent example demonstrates our 
approach.

\section{An Example}

We assumed a jointed body with rigid edge lengths $PQ=2$, $QR=3$, $RS=4$,
$SP=1$. Then randomly 19 positions of this body have been generated (see
fig.~\ref{figvier}) using the following degrees of freedom:
distance between $P$
and $R$, distance between $Q$ and $S$, rotations around X, Y and Z axes.
We measured
the distances between projected points in the respective frames which are
contained in  tab.~\ref{tabEins}.
(The high precision stems from the fact that we  simulated data.) 
Then we calculated the matrix of coefficients $f_{i,1}$..., $f_{i,19}$, $f_{i,
0}$ below for all frames $i$=1..19.
 given
in the Appendix B.

Then we solved for $x_1,....,x_{19}$ to obtain the result given below (for
most interesting variables):
\begin{eqnarray}
 x_1 = 4.00415, \, 
 x_2 = 8.98225, \, 
 x_3 = 15.9834, \, 
 x_4 = 0.999825.  \nn 
\end{eqnarray}
 (Original values were $a =   4, b=9, c=16, d=1$ resp.).

To investigate the impact of observational errors, the same experiment has
been repeated  assuming that the precision of measurement of position of
projected points is up to three leading digits. The results were: 
$$ x_1 = 3.88039, 
 x_2 = 8.55283, 
 x_3 = 15.1446, 
 x_4 = 0.976372. 
$$

\section{Discussion}

The example demonstrates the principal possibility of linear recovery of shape
parameters on a 4 part looped rigid object. Advantages and disadvantages  of
the approach are visible. The advantage is the need to solve a linear equation
system only instead of multivariable high order non-linear one. 
No restrictions are posed on the motion pattern of the object or on relative
motion of its jointed parts. The
disadvantages are: the great number of frames needed (nearly as much as used
by Rashid \cite{Rashid:1}),  and the danger of rounding  errors.

\unitlength=0.8cm
\input IN.PRO

A more close study should be devoted  to the impact of imprecision of
position of projections of traceable points. Experiments with reduction of
precision of up to three digits did not prove totally destructive.

One may wonder whether it is possible to apply the same technique for longer
loops of rigid parts. The eq.~(\ref{mostgeneral}) can be always "squared
out" to obtain an algebraic equation which in turn may be converted to a
linearized form. However, the number of variables  will explode making
practical application not feasible. We can see this already when comparing the
case of 3-part loops (see~\cite{Klopotek:92g}) and the 4-part loops (as
described above). Linearization in the former case increases the number of
necessary equations from 3 to 4, and in the latter case from 4 to 19.  

\begin{table}
\begin{center}
{\footnotesize
\begin{tabular}{rllll}
Frame                      &   $P'Q'$    &    $Q'R'$   &     $R'S'$  &
$S'P'$ \\
\hline
1           & 1.95661 &  1.44393 &  3.13125&  0.961803 \\
2           & 1.93014 &  2.91888 &  3.97348&  0.956746 \\
3           & 1.77619 &  1.90047 &   2.0017&  0.974922 \\
4           & 1.91128 &  1.42811 &  2.28367&  0.998392 \\
5           &  1.9462 &  2.92254 &  3.98388&  0.989842 \\
6           & 1.99945 &  2.97997 &  3.90664&  0.884827 \\
7           & 1.81095 &  1.96477 &  3.02693&  0.865462 \\
8           & 1.98808 &  2.58903 &  3.41279&  0.614607 \\
9           & 1.99891 &  2.11134 &   2.9322&  0.852155 \\
10          & 1.98413 &  2.76734 &  3.58316&  0.929779 \\
 11         & 1.80399 &  2.84376 &  3.97649&  0.940047 \\
  12        & 1.71086 &  2.55302 &  3.93766&  0.986488 \\
   13       & 1.75558 &  2.47493 &  3.85948&  0.949584 \\
    14      &  1.8613 &  1.78228 &  2.53998&   0.99849 \\
     15     & 1.99357 &  1.47441 &  2.84904&  0.999398 \\
      16    & 1.75743 &   2.2873 &  2.61136&  0.990972 \\
       17   & 1.66138 &  1.41837 &  2.12216&  0.930475 \\
18          & 1.99832 &  2.68278 &  3.63783&  0.915393 \\
19          & 1.84269 &  2.09825 &  1.97617&  0.831109 \\
\end{tabular}}
\end{center}
\caption{Distaces between projections of points $P,Q,R,S$ measured for frames
1-19 in a simulation run.} 
\label{tabEins} 
\end{table}

Still another question is the validity of the looped object model. Elsewhere
we demonstrate that for rigid bodies a test can be applied from frame to frame
to check whether or not the points belong to the same rigid body. Regrettably,
we cannot construct such a two-frame test for general looped objects. Only
after we have determined all the edge lengths, we can check whether or not
they fit geometrical requirements for each frame considered (whether they can
close a loop or not). If we want to guess which $n$    points constitute a
looped jointed object, we need to use some clues from the image,  e.g.
the fact that points are connected in the image by a straight line or by a
curved line.

We clearly can always run at risk of not being able to recover the structure
at all (the matrix of the equation system may be singular). This can
happen if the sequence of images is partially non-informative, e.g. 
 if the object does not move from frame to frame (so we get all
images identical), or if it is only shifted and not rotated, or if it is
rotated only around the axis perpendicular to the projection plane, etc. Also
the looped object may in fact behave like a rigid body, but in this case there
exists a simple test to detect this.

Last but not least we shall ask whether looped jointed objects may be observed
in
the reality. Though such objects are hardly observed in nature, many man-made
mechanisms possess a structure which operate like jointed looped objects
(think e.g. of the coupling system for Apollo space ships).

\section{Conclusion}
\begin{itemize}
\item A method has been outlined to recover 
 structural parameters of looped jointed objects from multiframes. 
\item This method has been linearized for 4-part loops. The number of
necessary frames is 19.
\item Extension of linearization approach to longer loops is possible in
principle, but will lead to unrealistic requirements concerning the number of
 necessary frames.
\end{itemize}

\newcommand{\A}[2]{ #2 #1}
\newcommand{\bibyear}[1]{}

\subsection*{Appendix A}
{\small Below we sketch transformations required to 'linearize' the equation
(\ref{exprX}):
%
\begin{equation} 
\sqrt{a-A_i} \pm \sqrt{b-B_i} \pm \sqrt{c-C_i} \pm \sqrt{d-D_i} = 0. \nn
\end{equation}

First we square the equation to obtain:
\begin{equation} 
(a-A_i)+(b-B_i)-(c-C_i)-(d-D_i)   \pm 2 \sqrt{(a-A_i) (b-B_i)} \pm 
2 \sqrt{(c-C_i) (d-D_i)} =0.
 \end{equation}
We square again to obtain:
\begin{eqnarray} 
&& (a-A_i)^2 + (b-B_i)^2 + (c-C_i)^2 + (d-D_i)^2 \nn\\
&&- 2   ( a - A_i )   ( b - B_i ) - 2   ( a - A_i )   (
c-C_i)-
2 (a-A_i) (d-D_i)-\nn\\
&&2 (b-B_i) (c-C_i)
-2 (b-B_i) (d-D_i)-2 (c-C_i) (d-D_i)\nn\\
&&+ 8 \sqrt{(a-A_i) (b-B_i) (c-C_i) (d-D_i)} =0.
\end{eqnarray}

And after the third squaring we finally eliminate square root and get:
\begin{eqnarray} 
\!\!\!\!\!\!\!\!\!\!\!\!&&+ (a-A_i)^4+ (b-B_i)^4+ (c-C_i)^4+ (d-D_i)^4 \nn\\
&&+6  (a-A_i)^2  (b-B_i)^2+6  (a-A_i)^2  (c-C_i)^2+6  (a-A_i)^2
  (d-D_i)^2 \nn\\
&&+6  (b-B_i)^2  (c-C_i)^2+6  (b-B_i)^2  (d-D_i)^2+6  (c-C_i)^2
  (d-D_i)^2 \nn\\
&&-40 (a-A_i) (b-B_i) (c-C_i) (d-D_i)\nn\\
&&-4  (a-A_i)^3 (b-B_i)-4  (a-A_i)^3 (c-C_i)-4  (a-A_i)^3 (d-D_i)\nn\\
  &&    -4 (a-A_i)  (b-B_i)^3-4  (b-B_i)^3 (c-C_i)-4  (b-B_i)^3 (d-D_i)\nn\\
&&-4 (a-A_i)  (c-C_i)^3-4 (b-B_i)  (c-C_i)^3-4  (c-C_i)^3 (d-D_i)\nn\\
  &&    -4 (a-A_i)  (d-D_i)^3-4 (b-B_i)  (d-D_i)^3-4 (c-C_i)  (d-D_i)^3\nn\\
&&+4  (a-A_i)^2 (b-B_i) (c-C_i)+4  (a-A_i)^2 (b-B_i) (d-D_i)\nn\\
&&+4  (a-A_i)^2
 (c-C_i) (d-D_i)\nn\\
&&+4 (a-A_i)  (b-B_i)^2 (c-C_i)+4 (a-A_i)  (b-B_i)^2 (d-D_i)+\nn\\
&&4  (b-B_i)^2
 (c-C_i) (d-D_i) +4 (a-A_i) (b-B_i)  (c-C_i)^2 \nn\\
  &&+4 (a-A_i)  (c-C_i)^2 (d-D_i)+4 (b-B_i)  (c-C_i)^2 (d-D_i)\nn\\
&& +4 (a-A_i) (b-B_i)  (d-D_i)^2+4 (a-A_i) (c-C_i)  (d-D_i)^2\nn\\
&&+4 (b-B_i) (c-C_i)  (d-D_i)^2   =0.
\end{eqnarray}

Finally, a rearrangement leads to equation (indices $i$ dropped):
\begin{eqnarray} 
&&({A}^4+{B}^4+{C}^4+{D}^4\nn\\
&&+6 {A}^2 {B}^2+6 {A}^2 {C}^2+6 {A}^2 {D}^2+6 {B}^2
{C}^2+6 {B}^2 {D}^2+6 {C}^2 {D}^2-40 A B C D\nn\\
&&-4 {A}^3 B-4 {A}^3 C-4 {A}^3 D-4 A {B}^3-4 {B}^3 C-4 {B}^3 D\nn\\
&&-4 A {C}^3-4 B {C}^3-4 {C}^3 D-4 A {D}^3-4 B {D}^3-4 C {D}^3\nn\\
&&+4 {A}^2 B C+4 {A}^2 B D+4 {A}^2 C D
+4 A {B}^2 C+4 A {B}^2 D+4 {B}^2 C D\nn\\
&&+4 A B {C}^2+4 A {C}^2 D+4 B {C}^2 D
+4 A B {D}^2+4 A C {D}^2+4 B C {D}^2) \nn\\
 &&+ a  (-4 {A}^3-12 A {B}^2-12 A {C}^2
-12 A {D}^2+40 B C D+4 {B}^3+4 {C}^3+4
{D}^3\nn\\
&&     +12 {A}^2 D+12 {A}^2 C+12 {A}^2 B
-8 A B C -8 A B D -8 A C D -4 {B}^2 C\nn\\
&& -4 C {D}^2 -4 B {C}^2 -4 {C}^2 D -4 B
{D}^2 -4 {B}^2 D )\nn\\
 &&+ b  (-4 {B}^3-12 B {C}^2-12 B {D}^2
-12 {A}^2 B+40 A C D+4 {A}^3+4 {C}^3+4
{D}^3\nn\\
&&     +12 {B}^2 D+12 {B}^2 C+12 A {B}^2
-8 A B C -8 A B D -8 B C D -4 {A}^2 C\nn\\
&& -4 {A}^2 D -4 A {C}^2 -4 {C}^2 D -4 A
{D}^2 -4 C {D}^2 )\nn\\
&& + c  (-4 {C}^3-12 C {D}^2-12 {A}^2 C
-12 {B}^2 C+40 A B D+4 {A}^3+4 {B}^3+4
{D}^3\nn\\
&&     +12 {C}^2 D+12 B {C}^2+12 A {C}^2
-8 A B C -8 A C D -8 B C D -4 {A}^2 B\nn\\
&& -4 {A}^2 D -4 A {B}^2 -4 {B}^2 D -4 A
{D}^2 -4 B {D}^2 )\nn\\
&& + d  (-4 {D}^3-12 {A}^2 D-12 {B}^2 D
-12 {C}^2 D+40 A B C+4 {A}^3+4 {B}^3+4
{C}^3\nn\\
&&     +12 C {D}^2+12 B {D}^2+12 A {D}^2
-8 A B D -8 A C D -8 B C D -4 {A}^2 B \nn\\
&&-4 {A}^2 C -4 A {B}^2 -4 {B}^2 C -4 A
{C}^2 -4 B {C}^2 )\nn\\
&& + {a}^2  (+6 {A}^2+6 {B}^2+6 {C}^2+6 {D}^2
-12 A B-12 A C-12 A D+4 B C+4 B D+4
C D )\nn\\
&& + {b}^2  (+6 {B}^2+6 {A}^2+6 {C}^2+6 {D}^2
-12 A B-12 B C-12 B D+4 A C+4 A D+4
C D )\nn\\
&& + {c}^2  (+6 {C}^2+6 {A}^2+6 {B}^2+6 {D}^2
-12 A C-12 B C-12 C D+4 A B+4 A D+4
B D )\nn\\
&& + {d}^2  (+6 {D}^2+6 {A}^2+6 {B}^2+6 {C}^2
-12 A D-12 B D-12 C D+4 A B+4 A C+4
B C )\nn\\
&&  + (a b)  (+24 A B-40 C D-12 {A}^2-12 {B}^2
+8 A C+8 A D+8 B C+8 B D+4 {C}^2+4
{D}^2 )\nn\\
&&  + (a c)  (+24 A C-40 B D-12 {A}^2-12 {C}^2
+8 A B+8 A D+4 {B}^2+8 B C+8 C D+4
{D}^2 )\nn\\
&&  + (a d)  (+24 A D-40 B C-12 {A}^2-12 {D}^2
+8 A B+8 A C+4 {B}^2+4 {C}^2+8 B
D+8 C D )\nn\\
&&  + (b c)  (+24 B C-40 A D-12 {B}^2-12 {C}^2
+4 {A}^2+8 A B+8 B D+8 A C+8 C D+4
{D}^2 )\nn\\
&&  + (b d)  (+24 B D-40 A C-12 {B}^2-12 {D}^2
+4 {A}^2+8 A B+8 B C+4 {C}^2+8 A
D+8 C D )\nn\\
&&  + (c d)  (+24 C D-40 A B-12 {C}^2-12 {D}^2
+4 {A}^2+4 {B}^2+8 A C+8 B C+8 A
D+8 B D )\nn\\
&&  + A  (-4 {a}^3-12 a {b}^2-12 a {c}^2
-12 a {d}^2+40 b c d+4 {b}^3+4 {c}^3+4
{d}^3
      +12 {a}^2 b +12 {a}^2 c +12 {a}^2 d\nn\\
&&      -4 {b}^2 c-4 {b}^2 d-4 b {c}^2
-4 {c}^2 d-4 b {d}^2-4 c {d}^2-8 a b c-8 a
b d-8 a c d )\nn\\
&&  + B  (-4 {b}^3-12 {a}^2 b-12 b {c}^2
-12 b {d}^2+40 a c d+4 {a}^3+4 {c}^3+4
{d}^3
      +12 a {b}^2 +12 {b}^2 c +12 {b}^2 d\nn\\
&&      -4 {a}^2 c-4 {a}^2 d-4 a {c}^2
-4 {c}^2 d-4 a {d}^2-4 c {d}^2-8 a b c-8 a
b d-8 b c d )\nn\\
&&  + C  (-4 {c}^3-12 {a}^2 c-12 {b}^2 c
-12 c {d}^2+40 a b d+4 {a}^3+4 {b}^3+4
{d}^3
      +12 a {c}^2 +12 b {c}^2 +12 {c}^2 d\nn\\
&&      -4 {a}^2 b-4 {a}^2 d-4 a {b}^2
-4 {b}^2 d-4 a {d}^2-4 b {d}^2-8 a b c-8 a
c d-8 b c d )\nn\\
&&  + D  (-4 {d}^3-12 {a}^2 d-12 {b}^2 d
-12 {c}^2 d+40 a b c+4 {a}^3+4 {b}^3+4
{c}^3
      +12 a {d}^2 +12 b {d}^2 +12 c {d}^2\nn\\
&&      -4 {a}^2 b-4 {a}^2 c-4 a {b}^2
-4 {b}^2 c-4 a {c}^2-4 b {c}^2-8 a b d-8 a
c d-8 b c d )\nn\\
&& +({a}^4+{b}^4+{c}^4+{d}^4
 +6 {a}^2 {b}^2+6 {a}^2 {c}^2+6 {a}^2 {d}^2
+6 {b}^2 {c}^2+6 {b}^2 {d}^2+6
{c}^2 {d}^2
 -40 a b c d  -4 {a}^3 b\nn\\
&&-4 {a}^3 c-4 {a}^3 d-4 a {b}^3-4 {b}^3 c-4 {b}^3 d
-4 a {c}^3-4 b {c}^3-4 {c}^3 d-4 a {d}^3-4 b {d}^3-4 c {d}^3 +4 {a}^2 b c\nn\\
&& +4{a}^2 b d+4 {a}^2 c d
+4 a {b}^2 c+4 a {b}^2 d+4 {b}^2 c d\nn\\
&& +4 a b {c}^2+4 a {c}^2 d+4 b {c}^2 d
+4 a b {d}^2+4 a c {d}^2+4 b c {d}^2 \label{mi}
  ).
\end{eqnarray}
Let us introduce variables:
\begin{eqnarray} 
&&x_1=a, \ \ \ x_2=b, \ \ \ x_3=c, \ \ \ x_4=d, 
x_5={a}^2, \ \ \ x_6={b}^2, \ \ \ x_7={c}^2, \ \ \ x_8={d}^2, \nn\\
&&x_9= (a b), \ \ \ x_{10}= (a c),  \ \ \ x_{11}=  (a d),  
 x_{12}=    (b c),\ \ \ x_{13}=  (b d),\ \ \ x_{14}=   (c d), \nn\\
&&x_{15}=(-4 {a}^3-12 a {b}^2-12 a {c}^2
-12 a {d}^2+40 b c d+4 {b}^3+4 {c}^3+4
{d}^3
      +12 {a}^2 b +12 {a}^2 c +12 {a}^2 d\nn\\
&&      -4 {b}^2 c-4 {b}^2 d-4 b {c}^2
-4 {c}^2 d-4 b {d}^2-4 c {d}^2-8 a b c-8 a
b d-8 a c d ),\nn\\
&&x_{16}=  (-4 {b}^3-12 {a}^2 b-12 b {c}^2
-12 b {d}^2+40 a c d+4 {a}^3+4 {c}^3+4
{d}^3
      +12 a {b}^2 +12 {b}^2 c +12 {b}^2 d\nn\\
&&      -4 {a}^2 c-4 {a}^2 d-4 a {c}^2
-4 {c}^2 d-4 a {d}^2-4 c {d}^2-8 a b c-8 a
b d-8 b c d ),\nn\\
&&x_{17}=  (-4 {c}^3-12 {a}^2 c-12 {b}^2 c
-12 c {d}^2+40 a b d+4 {a}^3+4 {b}^3+4
{d}^3
      +12 a {c}^2 +12 b {c}^2 +12 {c}^2 d\nn\\
&&      -4 {a}^2 b-4 {a}^2 d-4 a {b}^2
-4 {b}^2 d-4 a {d}^2-4 b {d}^2-8 a b c-8 a
c d-8 b c d ),\nn\\
&&x_{18}= (-4 {d}^3-12 {a}^2 d-12 {b}^2 d
-12 {c}^2 d+40 a b c+4 {a}^3+4 {b}^3+4
{c}^3
      +12 a {d}^2 +12 b {d}^2 +12 c {d}^2\nn\\
&&      -4 {a}^2 b-4 {a}^2 c-4 a {b}^2
-4 {b}^2 c-4 a {c}^2-4 b {c}^2-8 a b d-8 a
c d-8 b c d ),\nn\\
&&x_{19}=({a}^4+{b}^4+{c}^4+{d}^4
 +6 {a}^2 {b}^2+6 {a}^2 {c}^2+6 {a}^2 {d}^2
+6 {b}^2 {c}^2+6 {b}^2 {d}^2+6
{c}^2 {d}^2
 -40 a b c d\nn\\
&& -4 {a}^3 b-4 {a}^3 c-4 {a}^3 d-4 a {b}^3-4 {b}^3 c-4 {b}^3 d
 -4 a {c}^3-4 b {c}^3-4 {c}^3 d-4 a {d}^3-4 b {d}^3-4 c {d}^3\nn\\
&& +4 {a}^2 b c+4 {a}^2 b d+4 {a}^2 c d
+4 a {b}^2 c+4 a {b}^2 d+4 {b}^2 c d\nn\\
&& +4 a b {c}^2+4 a {c}^2 d+4 b {c}^2 d
+4 a b {d}^2+4 a c {d}^2+4 b c {d}^2 
  )
\end{eqnarray}
and constants (factors):
\begin{eqnarray} 
&&f_{i,0}=({A}^4+{B}^4+{C}^4+{D}^4
+6 {A}^2 {B}^2+6 {A}^2 {C}^2+6 {A}^2 {D}^2+6 {B}^2
{C}^2+6 {B}^2 {D}^2\nn\\
&&+6 {C}^2 {D}^2-40 A B C D
-4 {A}^3 B-4 {A}^3 C-4 {A}^3 D-4 A {B}^3-4 {B}^3 C-4 {B}^3 D
-4 A {C}^3\nn\\
&& -4 B {C}^3-4 {C}^3 D-4 A {D}^3-4 B {D}^3-4 C {D}^3
+4 {A}^2 B C+4 {A}^2 B D+4 {A}^2 C D
+4 A {B}^2 C \nn\\
&& +4 A {B}^2 D +4 {B}^2 C D
+4 A B {C}^2+4 A {C}^2 D+4 B {C}^2 D
+4 A B {D}^2+4 A C {D}^2+4 B C {D}^2), \nn\\
&&f_{i,1}=  (-4 {A}^3-12 A {B}^2-12 A {C}^2
-12 A {D}^2+40 B C D+4 {B}^3+4 {C}^3+4
{D}^3\nn\\
&&     +12 {A}^2 D+12 {A}^2 C+12 {A}^2 B
-8 A B C -8 A B D -8 A C D -4 {B}^2 C\nn\\
&& -4 C {D}^2 -4 B {C}^2 -4 {C}^2 D -4 B
{D}^2 -4 {B}^2 D ),\nn\\
&&f_{i,2}=  (-4 {B}^3-12 B {C}^2-12 B {D}^2
-12 {A}^2 B+40 A C D+4 {A}^3+4 {C}^3+4
{D}^3
     +12 {B}^2 D\nn\\
&&+12 {B}^2 C+12 A {B}^2
-8 A B C -8 A B D -8 B C D \nn\\
&&-4 {A}^2 C
 -4 {A}^2 D -4 A {C}^2 -4 {C}^2 D -4 A
{D}^2 -4 C {D}^2 ),\nn\\
&& f_{i,3}=  (-4 {C}^3-12 C {D}^2-12 {A}^2 C
-12 {B}^2 C+40 A B D+4 {A}^3+4 {B}^3+4
{D}^3
     +12 {C}^2 D\nn \\
&& +12 B {C}^2+12 A {C}^2
-8 A B C -8 A C D -8 B C D\nn\\  && -4 {A}^2 B
 -4 {A}^2 D -4 A {B}^2 -4 {B}^2 D -4 A
{D}^2 -4 B {D}^2 ),\nn\\
&& f_{i,4}=  (-4 {D}^3-12 {A}^2 D-12 {B}^2 D
-12 {C}^2 D+40 A B C+4 {A}^3+4 {B}^3+4
{C}^3\nn\\
&&     +12 C {D}^2+12 B {D}^2+12 A {D}^2
-8 A B D -8 A C D -8 B C D \nn \\ &&-4 {A}^2 B 
-4 {A}^2 C -4 A {B}^2 -4 {B}^2 C -4 A
{C}^2 -4 B {C}^2 ),\nn\\
&& f_{i,5}= (+6 {A}^2+6 {B}^2+6 {C}^2+6 {D}^2
-12 A B-12 A C-12 A D+4 B C+4 B D+4
C D ),\nn\\
&&f_{i,6}=  (+6 {B}^2+6 {A}^2+6 {C}^2+6 {D}^2
-12 A B-12 B C-12 B D+4 A C+4 A D+4
C D ),\nn\\
&& f_{i,7}=  (+6 {C}^2+6 {A}^2+6 {B}^2+6 {D}^2
-12 A C-12 B C-12 C D+4 A B+4 A D+4
B D )\nn\\
&& f_{i,8}= (+6 {D}^2+6 {A}^2+6 {B}^2+6 {C}^2
-12 A D-12 B D-12 C D+4 A B+4 A C+4
B C ),\nn\\
&& f_{i,9}=  (+24 A B-40 C D-12 {A}^2-12 {B}^2
+8 A C+8 A D+8 B C+8 B D+4 {C}^2+4
{D}^2 ),\nn\\
&& f_{i,10}=  (+24 A C-40 B D-12 {A}^2-12 {C}^2
+8 A B+8 A D+4 {B}^2+8 B C+8 C D+4
{D}^2 ),\nn\\
&&f_{i,11}=   (+24 A D-40 B C-12 {A}^2-12 {D}^2
+8 A B+8 A C+4 {B}^2+4 {C}^2+8 B
D+8 C D ),\nn\\
&& f_{i,12}=   (+24 B C-40 A D-12 {B}^2-12 {C}^2
+4 {A}^2+8 A B+8 B D+8 A C+8 C D+4
{D}^2 ),\nn\\
&&f_{i,13}=   (+24 B D-40 A C-12 {B}^2-12 {D}^2
+4 {A}^2+8 A B+8 B C+4 {C}^2+8 A
D+8 C D ),\nn\\
&&f_{i,14}=   (+24 C D-40 A B-12 {C}^2-12 {D}^2
+4 {A}^2+4 {B}^2+8 A C+8 B C+8 A
D+8 B D ),\nn\\
&&f_{i,15}=A, \ \ \ f_{i,16}=B, \ \ \ f_{i,17}=C, \ \ \ f_{i,18}=D, 
f_{i,19}=1.
\end{eqnarray}

Then, if we represent constant expressions as $f_{i,0}-f_{i,19}$ ($i$ - the
index of the frame), 
 we can write the above eq.~(\ref{mi}) as:

\newcommand{\VONFUENF}[5]{& #1}
\newcommand{\VONFUENX}[5]{\VONFUENF{#1}{#2}{#3}{#4}{#5} \\  }

\newcommand{\TEILEINS}{
\VONFUENF{ -407.58 }{ -23.3146 }{ 1006.6 }{ 2345.37 }{ 233.006 }
\VONFUENF{ -6299.13 }{ -4409.23 }{ 3592.69 }{ 14186.1 }{ 1518.94 }
\VONFUENF{ -365.78 }{ -266.823 }{ -160.799 }{ 661.211 }{ 2.17795 }
\VONFUENF{ -247.129 }{ -359.916 }{ 178.546 }{ 491.628 }{ -16.0673 }
\VONFUENF{ -6290.81 }{ -4512.79 }{ 3783.09 }{ 14225.6 }{ 1524.72 }
\VONFUENF{ -8257.51 }{ -4969.48 }{ 2930.08 }{ 15004.1 }{ 1392.25 }
\VONFUENX{ -1259.51 }{ -1187.45 }{ 965.538 }{ 3043.3 }{ 299.515 }
\VONFUENF{ -5305.18 }{ -3401.72 }{ 939.297 }{ 9075.91 }{ 629.862 }
\VONFUENF{ -1955.56 }{ -1787.29 }{ 605.637 }{ 3567.26 }{ 192.145 }
\VONFUENF{ -5749.96 }{ -3613.83 }{ 1911.79 }{ 9637.14 }{ 893.417 }
\VONFUENF{ -4177.35 }{ -3376.89 }{ 3389.65 }{ 12711.5 }{ 1588.96 }
\VONFUENF{ -705.908 }{ -1668.56 }{ 2962.61 }{ 10568.6 }{ 1436.68}
\VONFUENF{ -1201.6 }{ -1728.44 }{ 2732.44 }{ 10033.4 }{ 1248.24 }
\VONFUENX{ -671.446 }{ -712.076 }{ 374.275 }{ 1200.91 }{ 66.9038 }
\VONFUENF{ -574.514 }{ -478.336 }{ 778.026 }{ 1400.81 }{ 97.6985 }
\VONFUENF{ -1383.53 }{ -739.243 }{ 110.211 }{ 1890.16 }{ 213.264 }
\VONFUENF{ -190.213 }{ -227.683 }{ 137.781 }{ 340.682 }{ 10.5107 }
\VONFUENF{ -5327.86 }{ -3730.86 }{ 2269.42 }{ 10094.4 }{ 891.772 }
\VONFUENF{ -592.144 }{ -279.903 }{ -445.314 }{ 1078.64 }{ 4.91514 }
}


\newcommand{\TEILZWEI}{
\VONFUENF{  532.301 }{ -54.8179 }{ 785.302 }{ 496.262 }{ -156.663 }
\VONFUENF{ 237.559 }{ -302.159 }{ 2611.87 }{ 1782.89 }{ -291.517 }
\VONFUENF{ -34.0638 }{ -60.0113 }{ 270.887 }{ 181.345 }{ 171.155 }
\VONFUENF{ 144.302 }{ -91.9577 }{ 292.252 }{ 156.481 }{ 125.421 }
\VONFUENF{ 243.085 }{ -316.052 }{ 2621.5 }{ 1780.31 }{ -293.861 }
\VONFUENF{ 138.853 }{ -349.289 }{ 2634.06 }{ 1823.14 }{ 6.09405 }
\VONFUENX{ 207.413 }{ -134.336 }{ 826.781 }{ 625.601 }{ -10.2943 }
\VONFUENF{ 100.648 }{ -241.89 }{ 1679.41 }{ 1301.47 }{ 252.188 }
\VONFUENF{ 123.203 }{ -189.57 }{ 875.097 }{ 676.047 }{ 220.221 }
\VONFUENF{ 77.4807 }{ -320.514 }{ 1900.99 }{ 1323.33 }{ 165.563 }
\VONFUENF{ 297.994 }{ -213.503 }{ 2495.48 }{ 1678.93 }{ -545.265 }
\VONFUENF{ 489.941 }{ -70.8913 }{ 2125.17 }{ 1452.23 }{ -873.882 }
\VONFUENF{ 479.038 }{ -79.981 }{ 1981.56 }{ 1405.98 }{ -731.685 }
\VONFUENX{ 101.217 }{ -132.561 }{ 447.015 }{ 307.911 }{ 141.665 }
\VONFUENF{ 360.299 }{ -112.597 }{ 587.631 }{ 352.713 }{ 36.9919 }
\VONFUENF{ -54.2471 }{ -157.639 }{ 619.433 }{ 386.147 }{ 233.334 }
\VONFUENF{ 74.8051 }{ -69.7569 }{ 207.994 }{ 126.42 }{ 80.2594 }
\VONFUENF{ 170.395 }{ -296.226 }{ 1923.24 }{ 1396.35 }{ 51.5129 }
\VONFUENF{ -69.1469 }{ -36.6278 }{ 364.45 }{ 229.562 }{ 252.165 }
%
}
\newcommand{\TEILDREI}{
\VONFUENF{  -64.7885 }{ -342.622 }{ -746.125 }{ -681.854 }{ 3.82831 }
\VONFUENF{ -3285.47 }{ 190.944 }{ -521.13 }{ -1941.92 }{ 3.72543 }
\VONFUENF{ -270.569 }{ 171.824 }{ -225.21 }{ -207.536 }{ 3.15486 }
\VONFUENF{ -114.827 }{ 60.5441 }{ -397.509 }{ -158.64 }{ 3.65301 }
\VONFUENF{ -3286.55 }{ 199.117 }{ -528.368 }{ -1943.91 }{ 3.78771 }
\VONFUENF{ -3374.7 }{ 375.802 }{ -742.843 }{ -1903.37 }{ 3.99778 }
\VONFUENX{ -676.524 }{ 29.3802 }{ -480.488 }{ -687.047 }{ 3.27955 }
\VONFUENF{ -1918.18 }{ 345.7 }{ -832.739 }{ -1316.47 }{ 3.95246 }
\VONFUENF{ -793.006 }{ 237.396 }{ -659.423 }{ -682.11 }{ 3.99562 }
\VONFUENF{ -2364.95 }{ 394.058 }{ -710.478 }{ -1358.9 }{ 3.93679 }
\VONFUENF{ -3129.51 }{ 5.43114 }{ -270.197 }{ -1908.32 }{ 3.25437 }
\VONFUENF{ -2269.53 }{ -357.967 }{ -83.7823 }{ -1847.97 }{ 2.92705 }
\VONFUENF{ -1999.11 }{ -322.995 }{ -227.646 }{ -1753.4 }{ 3.08205 }
\VONFUENX{ -342.311 }{ 128.759 }{ -405.471 }{ -313.128 }{ 3.46446 }
\VONFUENF{ -120.19 }{ -106.042 }{ -673.335 }{ -423.168 }{ 3.97431 }
\VONFUENF{ -792.375 }{ 282.885 }{ -342.499 }{ -388.302 }{ 3.08854 }
\VONFUENF{ -119.161 }{ 52.3718 }{ -234.168 }{ -129.275 }{ 2.76018 }
\VONFUENF{ -2231.98 }{ 285.128 }{ -727.435 }{ -1462.76 }{ 3.99327 }
\VONFUENF{ -365.407 }{ 222.826 }{ -291.145 }{ -311.592 }{ 3.3955 }
%
}
%
\newcommand{\TEILVIER}{
\VONFUENF{  2.08493 }{ 9.80471 }{ 0.925065 }{ 1 }{ -2607.53 }
\VONFUENF{ 8.51988 }{ 15.7885 }{ 0.915362 }{ 1 }{ -2168.92 }
\VONFUENF{ 3.61179 }{ 4.00681 }{ 0.950472 }{ 1 }{ 533.38 }
\VONFUENF{ 2.03949 }{ 5.21517 }{ 0.996787 }{ 1 }{ 53.9041 }
\VONFUENF{ 8.54125 }{ 15.8713 }{ 0.979787 }{ 1 }{ -2902.02 }
\VONFUENF{ 8.88019 }{ 15.2619 }{ 0.782918 }{ 1 }{ 5169.05 }
\VONFUENX{ 3.86034 }{ 9.16229 }{ 0.749025 }{ 1 }{ -602.866 }
\VONFUENF{ 6.70309 }{ 11.6471 }{ 0.377742 }{ 1 }{ 7350.53 }
\VONFUENF{ 4.45775 }{ 8.59777 }{ 0.726168 }{ 1 }{ 1995.86 }
\VONFUENF{ 7.65817 }{ 12.839 }{ 0.864489 }{ 1 }{ 4358.75 }
\VONFUENF{ 8.08695 }{ 15.8124 }{ 0.883689 }{ 1 }{ -5982.04 }
\VONFUENF{ 6.51791 }{ 15.5051 }{ 0.973158 }{ 1 }{ -10819.7 }
\VONFUENF{ 6.12529 }{ 14.8956 }{ 0.901709 }{ 1 }{ -8864.45 }
\VONFUENX{ 3.17654 }{ 6.45152 }{ 0.996982 }{ 1 }{ 244.05 }
\VONFUENF{ 2.17387 }{ 8.11704 }{ 0.998796 }{ 1 }{ -1097.81 }
\VONFUENF{ 5.23172 }{ 6.81921 }{ 0.982025 }{ 1 }{ 1383.22 }
\VONFUENF{ 2.01179 }{ 4.50357 }{ 0.865783 }{ 1 }{ 16.9022 }
\VONFUENF{ 7.19729 }{ 13.2338 }{ 0.837944 }{ 1 }{ 2409.01 }
\VONFUENF{ 4.40265 }{ 3.90527 }{ 0.690742 }{ 1 }{ 1059.24 }
}

\begin{table}
\begin{center}
{\scriptsize
 \begin{tabular}{rrrrrrrr}
 \renewcommand{\VONFUENF}[5]{& #1} 
 $f_1$ \TEILEINS \\
 \renewcommand{\VONFUENF}[5]{& #2} 
 $f_2$ \TEILEINS \\
 \renewcommand{\VONFUENF}[5]{& #3}
 $f_3$ \TEILEINS \\
 \renewcommand{\VONFUENF}[5]{& #4}
 $f_4$ \TEILEINS \\
 \renewcommand{\VONFUENF}[5]{& #5}
 $f_5$ \TEILEINS \\
 \renewcommand{\VONFUENF}[5]{& #1} 
 $f_6$ \TEILZWEI \\
 \renewcommand{\VONFUENF}[5]{& #2} 
 $f_7$ \TEILZWEI \\
 \renewcommand{\VONFUENF}[5]{& #3}
 $f_8$ \TEILZWEI \\
 \renewcommand{\VONFUENF}[5]{& #4}
 $f_9$ \TEILZWEI \\
 \renewcommand{\VONFUENF}[5]{& #5}
 $f_{10}$ \TEILZWEI \\
 \renewcommand{\VONFUENF}[5]{& #1} 
 $f_{11}$ \TEILDREI \\
 \renewcommand{\VONFUENF}[5]{& #2} 
 $f_{12}$ \TEILDREI \\
 \renewcommand{\VONFUENF}[5]{& #3}
 $f_{13}$ \TEILDREI \\
 \renewcommand{\VONFUENF}[5]{& #4}
 $f_{14}$ \TEILDREI \\
 \renewcommand{\VONFUENF}[5]{& #5}
 $f_{15}$ \TEILDREI \\
 \renewcommand{\VONFUENF}[5]{& #1} 
 $f_{16}$ \TEILVIER \\
 \renewcommand{\VONFUENF}[5]{& #2} 
 $f_{17}$ \TEILVIER \\
 \renewcommand{\VONFUENF}[5]{& #3}
 $f_{18}$ \TEILVIER \\
 \renewcommand{\VONFUENF}[5]{& #4}
 $f_{19}$ \TEILVIER \\
 \renewcommand{\VONFUENF}[5]{& #5}
 $f_{0}$ \TEILVIER \\
 \end{tabular} }
\end{center}
\caption{Factors $f_0-f_{19}$ for the frames from the example
from section 3 
} \label{tabZwei}
\end{table}
\begin{eqnarray} 
&&f_{i,1} x_1 + f_{i,2} x_2 + f_{i,3} x_3 + f_{i,4} x_4 + f_{i,5} x_5 
+f_{i,6} x_6 + f_{i,7} x_7 + f_{i,8} x_8 + f_{i,9} x_9 + f_{i,10} x_{10}
\nn\\ 
&&+f_{i,11} x_{11}+f_{i,12} x_{12}+f_{i,13} x_{13}+f_{i,14} x_{14} 
+f_{i,15} x_{15} \nn\\
&&+f_{i,16}x_{16}+f_{i,17}x_{17}+f_{i,18}x_{18}+f_{i,19}x_{19}+ f_{i,0} =0
\end{eqnarray}
}

\subsection*{Appendix B}

{\small In Table \ref{tabZwei} we give coefficients of equations obtained for
our example. 
%
%
%
%

The solution was 
\begin{eqnarray}
&& x1 = 4.00415, 
 x2 = 8.98225, 
 x3 = 15.9834, 
 x4 = 0.999825, \nn\\
 && x5 = 9.36547, 
 x6 = 73.8661, 
 x7 = 248.763, 
 x8 = -5.91918, 
 x9 = 29.1833,   \nn\\
 &&x10 = 57.1582, 
 x11 = -2.78924, 
 x12 = 136.854, 
 x13 = 2.10081, 
 x14 = 9.15377,   \nn\\
 &&x15 = -7655.73, 
 x16 = -5115.85, 
 x17 = 3834.1, 
 x18 = 15333.2, 
 x19 = -13.2689
\end{eqnarray}
}

\bigskip
\begin{footnotesize}
\begin{minipage}[c]{30mm}

\end{minipage}
\begin{minipage}[c]{98mm}
{\bf Mieczys{\l}aw A. K{\l}opotek } 
was born in Brusy, 
Poland, 
in 1960. He received his Master's (in 1983) and Ph.D. (in 1984) degrees in 
Information 
Processing from the University of Technology in Dresden, Germany. Since 1985 
he works in the Institute of Computer Science of Polish Academy of Sciences in
 Warsaw. His research interests are in computer vision, artificial 
intelligence and creative data analysis.
\end{minipage}
\end{footnotesize}
\end{document}